\documentclass[10pt,twocolumn,letterpaper]{article}

\usepackage{cvpr}              

\usepackage{graphicx}
\usepackage{amsmath}
\usepackage{amssymb}
\usepackage{booktabs}
\usepackage{tabularx}
\usepackage{bm}
\usepackage{cite}
\usepackage{float}
\usepackage{wrapfig}
\usepackage{calc}
\usepackage[svgnames,x11names,table]{xcolor}
\usepackage{multirow}
\usepackage{tikz}
\usepackage{mdframed}
\usetikzlibrary{positioning}
\usepackage{times}
\usepackage{latexsym}
\usepackage[T1]{fontenc}
\usepackage[utf8]{inputenc}
\usepackage{microtype}
\usepackage{inconsolata}
\usepackage{listings}
\usepackage{here}
\usepackage{mathtools}
\usepackage{ulem}
\usepackage{soul}
\usepackage {colortbl,array}
\usepackage{enumitem}
\usepackage{xspace}

\definecolor{TitleColor}{gray}{0.95}
\definecolor{LightCyan}{rgb}{0.88,0.95,1}
\definecolor{LightPink}{HTML}{FAE6E7}
\definecolor{MintGreen}{HTML}{cbe7d1}

\definecolor{att}{HTML}{A8DDA8}      
\definecolor{rel}{HTML}{BFEFFF}      
\definecolor{obj}{HTML}{FFAEB9}      
\definecolor{num}{HTML}{FFD700}      
\definecolor{txt}{HTML}{D8BFD8}      
\definecolor{fact}{HTML}{98FB98}     

\usepackage[most]{tcolorbox}
\usepackage{cuted}
\tcbuselibrary{breakable}

\newcommand{\WideTColorBox}[2]{%
  \par\begingroup
  \onecolumn
  \begin{tcolorbox}[enhanced, breakable,
    width=\textwidth,
    colback=blue!5!white,
    colframe=blue!75!black,
    fonttitle=\bfseries,
    title=#1,
    boxrule=0.5pt,
    sharp corners,
    left=2mm, right=2mm, top=1mm, bottom=1mm,
    before upper={\let\\\par},
  ]
  #2
  \end{tcolorbox}
  \endgroup
  \par
}

\newcommand{\TColorBox}[2]{%
  \begin{tcolorbox}[enhanced, breakable,
    colback=blue!5!white,
    colframe=blue!75!black,
    fonttitle=\bfseries,
    title=#1,
    boxrule=0.5pt,
    sharp corners,
    left=2mm, right=2mm, top=1mm, bottom=1mm,
    width=\columnwidth,
    before upper={\let\\\par},
  ]
  #2
  \end{tcolorbox}
}

\newcommand{\modelname}{ReMoRa}

\usepackage[pagebackref,breaklinks,colorlinks,citecolor=Blue4]{hyperref}

\usepackage[capitalize]{cleveref}
\crefname{section}{Sec.}{Secs.}
\Crefname{section}{Section}{Sections}
\Crefname{table}{Table}{Tables}
\crefname{table}{Tab.}{Tabs.}

\def\confName{CVPR}
\def\confYear{2026}

\begin{document}

\title{ReMoRa: Multimodal Large Language Model based on \\ Refined Motion Representation for Long-Video Understanding}

\author{
  Daichi Yashima$^{1,3}$ \quad Shuhei Kurita$^{2,3}$ \quad Yusuke Oda$^{3}$ \quad Komei Sugiura$^{1}$ \\
  $^{1}$Keio University \quad $^{2}$NII \quad $^{3}$NII LLMC \\
}

\maketitle
\vspace{-3mm}
\begin{abstract}
While multimodal large language models (MLLMs) have shown remarkable success across a wide range of tasks, long-form video understanding remains a significant challenge.
In this study, we focus on video understanding by MLLMs.
This task is challenging because processing a full stream of RGB frames is computationally intractable and highly redundant, as self-attention have quadratic complexity with sequence length.
In this paper, we propose ReMoRa, a video MLLM that processes videos by operating directly on their compressed representations.
A sparse set of RGB keyframes is retained for appearance, while temporal dynamics are encoded as a motion representation, removing the need for sequential RGB frames.
These motion representations act as a compact proxy for optical flow, capturing temporal dynamics without full frame decoding.
To refine the noise and low fidelity of block-based motions, we introduce a module to denoise and generate a fine-grained motion representation.
Furthermore, our model compresses these features in a way that scales linearly with sequence length.
We demonstrate the effectiveness of ReMoRa through extensive experiments across a comprehensive suite of long-video understanding benchmarks.
ReMoRa outperformed baseline methods on multiple challenging benchmarks, including LongVideoBench, NExT-QA, and MLVU.

\end{abstract}

\section{Introduction}
\begin{figure}[t]
    \centering
    \includegraphics[width=1.0\linewidth]{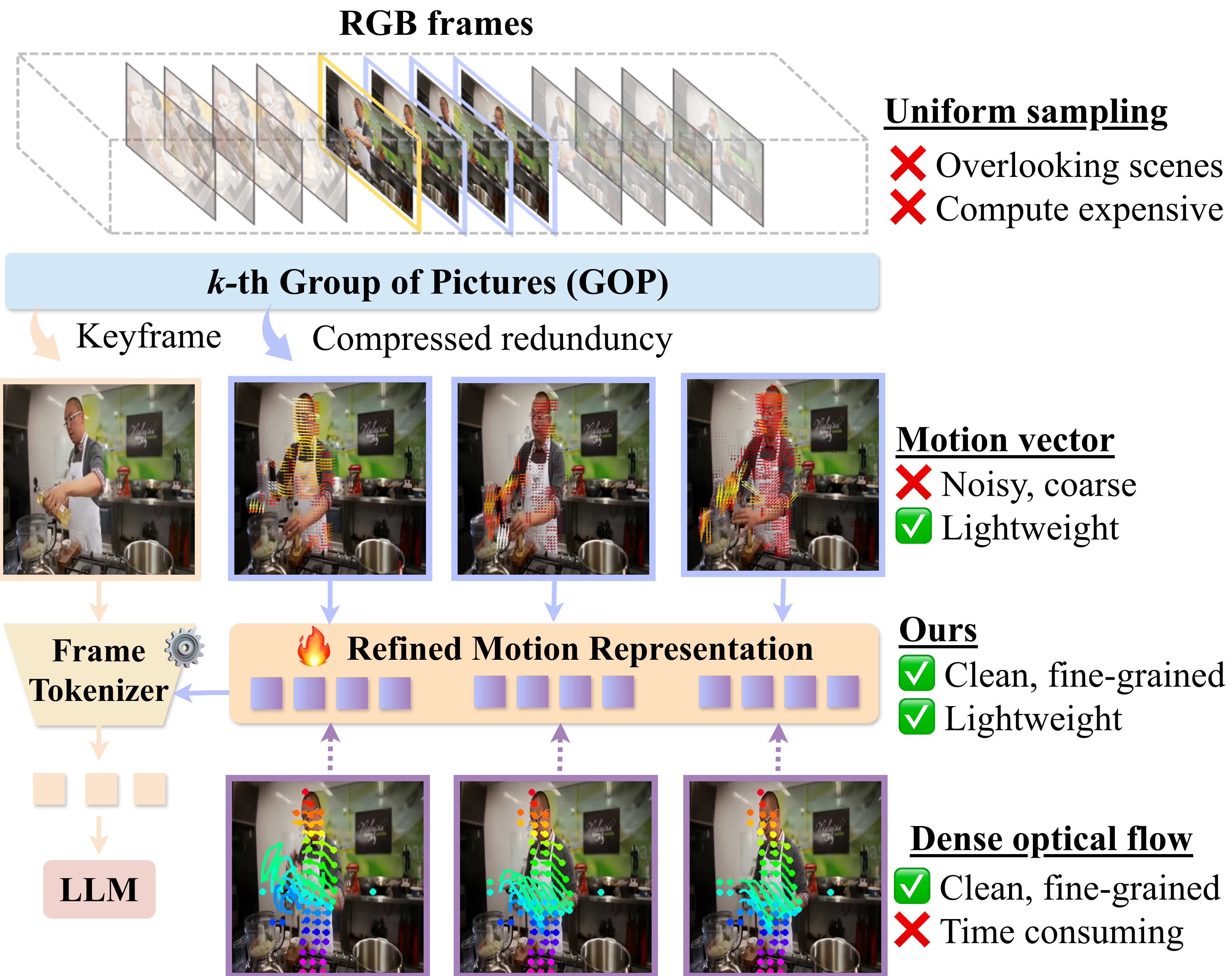}
    \caption{\textbf{Overview of ReMoRa.} 
    Our method utilizes compressed video representations, which naturally separates each video into keyframes and compressed inter-frame redundancies. 
    From these, we extract motions that are lightweight but noisy and coarse. Our model then refines these motions into clean and fine-grained representations that preserve efficiency while approaching the fidelity of dense optical flow. 
    }
    \label{fig:eye-catch}
    \vspace{-3mm}
\end{figure}
Multimodal large language models (MLLMs) have demonstrated remarkable capabilities across a wide range of vision-language tasks~\cite{gpt4v, reid2024gemini, llavaonevision, flamingo, chen2024internvl, blip2, llava, yang2024arxiv, zhao2025cobra}.
However, this success has not yet translated to the complex domain of long-form videos, which remains a significant open challenge~\cite{longva, vamba, llavaonevision, maaz24acl}.
Current MLLMs often struggle when confronted with inputs spanning several minutes or even hours, leading to a degradation in performance.
This bottleneck severely limits their practical utility in real-world scenarios that demand the comprehension of complex, long-range temporal narratives, such as video summarization~\cite{videocaptioningsurvey}, visual scene interpretation for domestic robots~\cite{relaxformer, pi0}, or assistive technologies for the visually impaired~\cite{tang2023video, nguyen-etal-2024-video}.

The primary obstacle to effective long-video understanding is the cost and redundancy of processing long RGB frame sequences.
Most existing models rely on uniform frame sampling, which entails an inherent trade-off~\cite{llavaonevision, shen2025longvu, Islam_2025_CVPR, llavavideo, videollava, llavanextvideo}:
sparse sampling is computationally feasible but misses brief yet critical events, whereas dense sampling preserves fine-grained dynamics but quickly becomes prohibitive due to the quadratic complexity of self-attention on long sequences.
Moreover, frame-based approaches are inefficient, repeatedly encoding redundant visual content (e.g., static backgrounds) across consecutive frames.
Recent methods attempt to mitigate this by maintaining dense uniform sampling and then compressing tokens in time or space~\cite{videollava, llavavideo}, but pooling- and reduction-based strategies often blur fine-grained detail and subtle motion cues, leading to underperformance on tasks that depend on short-lived but salient events, especially as video length grows~\cite{shen2025longvu, Islam_2025_CVPR}.
These limitations suggest that long-form video MLLMs should first adopt a more compact, redundancy-aware video representation before applying heavy sequence modeling.

A natural candidate for such a representation is the compressed video stream itself.
Modern video compression formats (e.g., H.264) factorize a video into a small number of keyframes and many lightweight motion descriptors that describe scene evolution over time.
Operating in this space both avoids repeated processing of nearly identical frames and enables much denser temporal coverage for a fixed computational budget by using inexpensive motion information instead of fully decoded RGB frames.

In this paper, we realize this compressed-domain perspective with a standard video codec while keeping the method independent of any specific format.
Standard codecs organize frames into {groups of pictures} (GOPs), where each GOP consists of an I-frame and several P/B-frames predicted from one or more reference frames.
This structure induces a temporal hierarchy: I-frames serve as appearance anchors, while P/B-frames capture local motion between anchors.
Concretely, our model uses I-frames as keyframes and motion vectors from P/B-frames as motion descriptors, a computationally efficient proxy for optical flow~\cite{coviar, slowifastp, mfcd_net} that preserves appearance from a small set of RGB frames while capturing temporal dynamics through lightweight motion vectors.
This efficiency comes with trade-offs, as motion vectors in standard codecs are block-based, sparse, and predictive rather than directly observed. The resulting signals are noisy and temporally inconsistent, undermining temporal alignment and fine-grained reasoning. Crucially, their very low computational cost permits sampling more temporal positions than typical RGB-based schemes. Taken together, these factors motivate transforming coarse, codec-derived motions into fine-grained motion representations, enabling dense temporal coverage and motion reasoning at a fraction of the cost of full optical flow.

Motivated by these considerations, we propose ReMoRa, a video MLLM that exploits compressed video representations for long video understanding. 
As shown in Fig.~\ref{fig:eye-catch}, instead of uniformly sampling frames, we keep a small set of high-information RGB frames for appearance and encode temporal change as motion representations obtained from compressed-domain motion cues, a computationally efficient proxy for optical flow. 
Subsequently, ReMoRa employs a Refined Motion Representation (RMR) module that denoises and densifies block-level raw motions into fine-grained motion representations aligned with dense optical flow that are suitable for long-horizon reasoning.
Furthermore, to efficiently capture temporal dependencies, we introduce the Hierarchical Motion State Space (HMSS) module that exploits the codec-induced structure to compress motion sequences in linear time, avoiding quadratic self-attention while preserving both local and global temporal context.
We validated the effectiveness of our approach on a comprehensive suite of video understanding benchmarks such as VideoMME~\cite{videomme}, LongVideoBench~\cite{longvideobench}, and ActivityNet-QA~\cite{activitynetqa}.

Our contributions are summarized as follows:
\begin{itemize}
    \setlength{\parskip}{0.2mm} 
    \setlength{\itemsep}{0.2mm} 
    \item We present ReMoRa, a video MLLM that directly processes compressed video streams, leveraging I-frames and motion representations instead of redundant RGB frames for scalable long-form video understanding.
    \item We propose the RMR module to enhance the fidelity of noisy, block-level motion vectors and introduce a HMSS module to perform long-range temporal modeling in linear time.
    \item ReMoRa outperformed baseline methods on multiple challenging long-form video understanding benchmarks, including LongVideoBench~\cite{longvideobench}, NExT-QA~\cite{nextqa}, and MLVU~\cite{mlvu}.
\end{itemize}

\section{Related Work}
\begin{figure*}[t]
    \centering
    \includegraphics[width=1.0\linewidth]{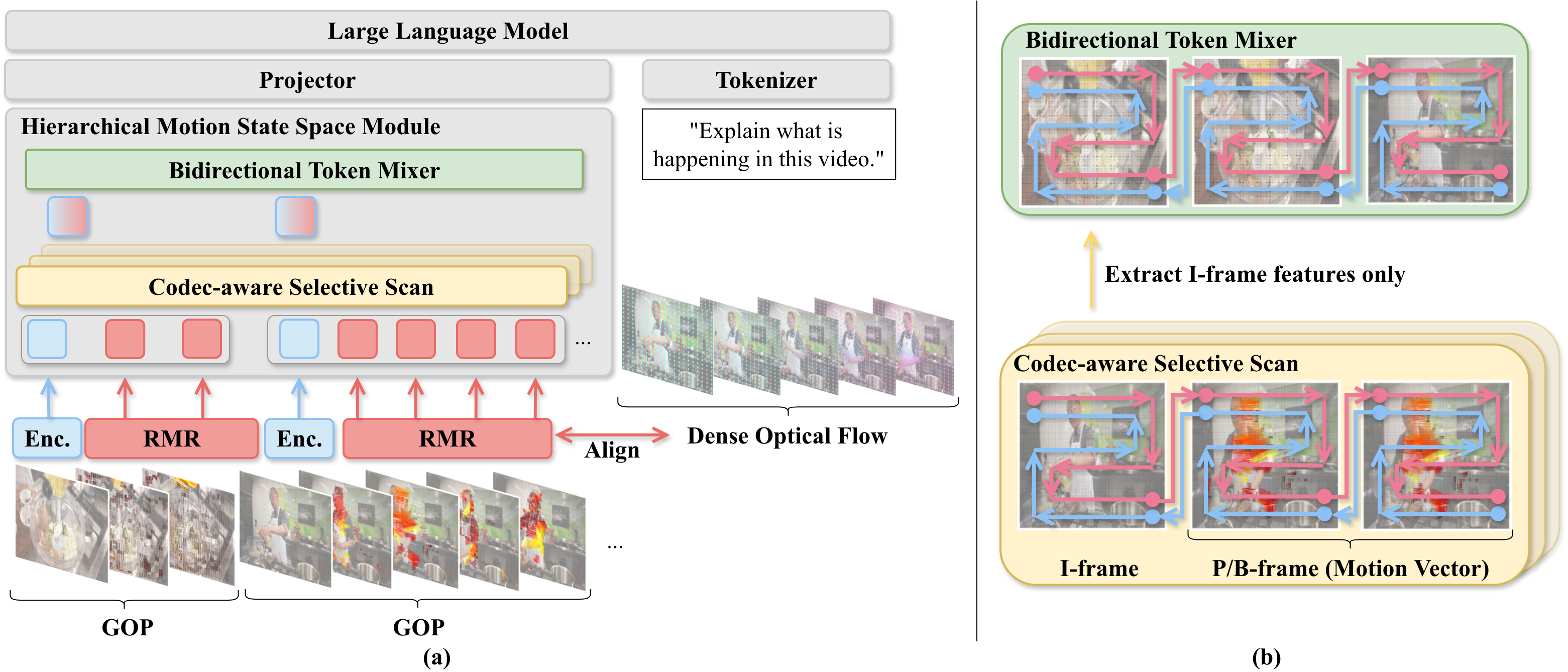}
    \caption{\textbf{Architecture of ReMoRa}: The model operates directly in the compressed video representation for long-video understanding. (a) It consists of an image encoder, the Refined Motion Representation (RMR) module, the Hierarchical Motion State Space (HMSS) Module, and a pretrained LLM.
     Each clip is decomposed into group of pictures (GOPs) with a single I-frame and several P/B frames represented by motion vectors. 
     The image encoder (Enc.) extracts patch embeddings from I-frames, while the RMR module converts coarse motion vectors into dense, high-fidelity representations. 
    (b) The HMSS module fuses the refined motions and appearance features within each GOP and models long-range dependencies across GOPs through a state space model, enabling linear-time temporal reasoning before alignment with the LLM.} 
    \label{fig:model}
    \vspace{-4mm}
\end{figure*}

Recent studies have explored various aspects of video MLLMs, including their architectures, training paradigms, and for effective modeling their unique visual and temporal dynamics~\cite{liang2024survey, li2024multimodal}.

\paragraph{Video Multimodal Large Language Models.}
Recent video MLLMs extend image-based MLLMs such as LLaVA~\cite{llava} and MiniGPT-4~\cite{zhu2024minigpt} to the temporal domain, yet they face a severe frame budget constraint due to quadratic attention and limited context windows. 
Early models~\cite{maaz24acl, videollama, llavaonevision} processed only a handful of uniformly sampled RGB frames (e.g., 8$\sim$32) to align video content with language, but this approach discards fine-grained temporal cues and fails to scale to minute- or hour-long videos.
More advanced models (e.g.,~\cite{damonlpsg2025videollama3, chen2025longvila, nan2025openvidm}) introduce sparse or segment-based sampling, yet they still rely on frame-level decoding.

To overcome the inefficiency of uniform frame sampling, several work have proposed token-efficient mechanisms. 
LongVU~\cite{shen2025longvu} dynamically prunes redundant visual tokens to fit long clips within a manageable context.
BIMBA~\cite{Islam_2025_CVPR} employs  state space models (SSMs)~\cite{gu22iclr,gu2024mamba} to model long temporal dependencies linearly in sequence length, compressing frame representations into compact temporal states.
Despite such efficiency gains, these approaches still process decoded RGB frames and consequently suffer from redundant background information and heavy preprocessing costs.

\paragraph{Compressed Video Understanding.}
An emerging research direction leverages the intrinsic compression structures within video streams to enable efficient visual representation learning~\cite{babu2016survey, Madhushree2019survey}.
 Modern codecs such as H.264/AVC and HEVC~\cite{hevc} inherently perform keyframe selection and motion compensation through the group of pictures (GOPs) structure, providing a natural appearance--motion decomposition. 
This insight has inspired work in compressed-domain video understanding.
CoViAR~\cite{coviar} first demonstrated that by using I-frames, motion vectors, and residuals, models can achieve competitive action recognition without full decoding.
Follow-up models~\cite{dmc_net, mfcd_net} refine motion vectors to approximate dense optical flow, improving motion fidelity under high compression.
\vspace{-0.5mm}

Building upon these findings, recent video MLLMs integrate codec-derived motion cues directly into multimodal reasoning.
Video-LaVIT~\cite{video-lavit} decomposes videos into appearance tokens from I-frames and motion tokens from motion vectors for video-language alignment.
EMA~\cite{zhao25cvpr} introduces a motion-aware GOP encoder that fuses sparse RGB keyframes with compressed motion cues to achieve efficient temporal reasoning.
Our approach follows this codec-aware paradigm but further refines block-level motion vectors through the RMR module, recovering dense optical-flow motion without expensive decoding.
Furthermore, we propose the HMSS module that mirrors the GOP hierarchy, enabling linear-time modeling of long-range dependencies across the compressed sequence.

\section{Preliminaries}
\label{sec:preliminaries}

\label{sec:codec}

In modern video compression standards such as H.264/AVC and HEVC~\cite{hevc}, intra-coded frames (I-frames) and predictive/bipredictive-coded frames (P/B-frames) play complementary roles in reducing temporal redundancy.
I-frames are self-contained and encoded without reference to other frames, serving as reliable anchor points for random access, error recovery, and scene changes.
In contrast, P-frames are encoded by predicting their content from previously decoded reference frames using motion estimation and motion compensation. 
Similarly, B-frames are encoded using motion and residual information from both preceding and succeeding reference frames.
Instead of storing full pixel data, P/B-frames record motion vectors and residual differences relative to the reference frame, significantly lowering the amount of transmitted data.
By alternating I-frames and P/B-frames, encoders achieve an efficient balance between compression ratio and decoding stability, maintaining both visual quality and temporal coherence.

Formally, let a video be denoted as
$\mathbf{V} \in \mathbb{R}^{T \times H \times W \times 3}$ where $T$, $H$, and $W$ denote the number of frames, height, and width, respectively.
Standard video codecs (e.g., H.264) segment $\mathbf{V}$ into $K$ GOPs. 
Each $\text{GOP}^{(k)}$ ($k \in \{0, \ldots, K-1\}$)
consists of one I-frame, $\mathbf{V}^{(k,0)}$, and a subsequent set of P/B-frames, $\{\mathbf{V}^{(k,t)}\}_{t=1}^{T_k-1}$.
The length $T_k$ of GOPs in a raw compressed stream is often variable, depending on scene complexity and encoder settings.
To create uniform tensor shapes, we pad all GOPs to a fixed maximum length, denoted $T_g$.
Thus, $T_g$ represents the standardized temporal dimension for all GOPs processed by our model.

The encoding of each inter-coded frame $\mathbf{V}^{(k,t)}$ ($t>0$) relies on motion compensation. The frame is partitioned into blocks $\mathbf{P}^{(k,t)}(u,v) \in \mathbb{R}^{b_h \times b_w \times 3}$, where $(u,v)$, $b_h$, and $b_w$ are the indices, height, and width of the block, respectively.
For each block, the encoder searches a reference frame $\mathbf{V}^{(k',t')}$ for the most similar block and stores the displacement as a motion vector:
{
\setlength{\abovedisplayskip}{6pt}
\setlength{\belowdisplayskip}{6pt}
\setlength{\abovedisplayshortskip}{6pt}
\setlength{\belowdisplayshortskip}{6pt}
\begin{align}
\mathbf{m}^{(k,t)}(u,v) = \mathbf{P}^{(k', t')}(u', v') - \mathbf{P}^{(k, t)}(u, v)
\end{align}
}
The predicted block $\hat{\mathbf{P}}^{(k,t)}(u,v)$ is formed from the reference block at the displaced location.
The codec then encodes only the residual between the true and predicted frames,
$\mathbf{R}^{(k,t)} = \mathbf{V}^{(k,t)} - \hat{\mathbf{V}}^{(k,t)}$,
together with the motion field $\mathbf{m}^{(k,t)}$, which is the collection of all block-level vectors $\mathbf{m}^{(k,t)} = [\mathbf{m}^{(k,t)}(1,1), \dots, \mathbf{m}^{(k,t)}(W/b_w, H/b_h)]$.
At decoding, the P/B-frame is reconstructed in pixel space as:
{
\setlength{\abovedisplayskip}{6pt}
\setlength{\belowdisplayskip}{6pt}
\setlength{\abovedisplayshortskip}{6pt}
\setlength{\belowdisplayshortskip}{6pt}
\begin{align}
\tilde{\mathbf{V}}^{(k,t)} = \mathbf{f}(\mathbf{V}^{(k',t')}, \mathbf{m}^{(k,t)}) + \tilde{\mathbf{R}}^{(k,t)},
\end{align}
}
where $\mathbf{f}(\cdot)$ is the motion compensation function and $\tilde{\mathbf{R}}^{(k,t)}$ is the decoded residual.

From a vision perspective, the motion field $\mathbf{m}^{(k,t)}$ can be regarded as a block-level approximation to optical flow.
It provides compact, codec-native motion cues that capture temporal dynamics at low computational cost
and has proven effective in compressed-domain video understanding~\cite{video-lavit, zhao25cvpr}.
However, since it is estimated through block-based matching within the codec pipeline, it is inherently noisy, spatially coarse, and temporally sparse, often spanning long intervals between keyframes.
Such limitations can hinder fine-grained motion reasoning and temporal consistency.

\section{Method}
\label{sec:method}

We propose \modelname, a video MLLM that directly utilizes compressed video representations as inputs to model long-context video while preserving high fidelity motion dynamics.
Fig.~\ref{fig:model} shows the overall architecture of \modelname. 
The model consists of an image encoder~\cite{zhai23iccv}, the Refined Motion Representation (RMR) module, the Hierarchical Motion State Space (HMSS) module and a pretrained LLM~\cite{yang2024arxiv}.

First, each clip is decomposed into GOP tuples containing RGB information from I-frames and their subsequent P/B-frames (i.e., motion vectors).
Then, an image encoder extracts spatial patch embeddings of the I-frame and the RMR module is used to obtain refined dense motion representations from the P/B-frames.
Finally, the HMSS module, a SSM with linear complexity in sequence length, fuses the refined motion vectors with the I-frame features and outputs a temporally aggregated representation for the subsequent LLM.
By structuring its computation around GOP tuples, ReMoRa inherently preserves local temporal coherence and mitigates redundancy, yielding a more effective representation for aligning dynamic video content with textual prompts.

\paragraph{Compressed Video Representation-aware Input.}
Our model receives the tokenized instruction sequence $\bm{x}_{\text{txt}} \in \mathbb{R}^{V \times L}$ and the video input, which is structured as a list of $K$ GOPs: $[\text{GOP}^{(0)}, \text{GOP}^{(1)}, \ldots, \text{GOP}^{(K-1)}]$. 
Here, $V$ and $L$ are the vocabulary size and the textual sequence length, respectively.
We define the input for $\text{GOP}^{(k)}$ as a tuple:
{
\setlength{\abovedisplayskip}{6pt}
\setlength{\belowdisplayskip}{6pt}
\setlength{\abovedisplayshortskip}{6pt}
\setlength{\belowdisplayshortskip}{6pt}
\begin{align}
\label{eq:gop_tuple}
\text{GOP}^{(k)} = \left( \mathbf{V}^{(k, 0)}, \mathbf{m}^{(k, 1)}, \ldots, \mathbf{m}^{(k, T_g)} \right).
\end{align}
}
Each tuple $\text{GOP}^{(k)}$ begins with one I-frame $\mathbf{V}^{(k, 0)} $,
followed by $T_g-1$ P/B-frames, which are represented only by their blockwise motion-vector fields
$\mathbf{m}^{(k, t)} $.
This representation enables the modeling of temporal dynamics directly in the compressed domain, without reconstructing pixel-level frames.

Each I-frame $\mathbf{V}^{(k, t)}$ is independently processed as follows:
First, it is divided into $N_\text{p}$ non-overlapping patches, each with size $p\times p$, where $N_\text{p} = (H\times W) / p^2$.
The patches are then input to a vision encoder~\cite{zhai23iccv} to obtain the I-frame patch embeddings
$\bm{E}_{\text{I}}^{(k)} \in \mathbb{R}^{N_\text{p} \times d_\text{s}}$, where $d_\text{s}$ is the embedding dimension.

\subsection{Refined Motion Representation Module}
\label{sec:refiner}
We propose the RMR module, that enhances sparse motions into fine-grained representations, capturing richer dynamics while mitigating compression artifacts.
Raw motions offer a significant computational advantage over processing fully decoded inter-coded frames, as the inherent sparsity of motion vectors allows us to naturally discard redundant information from static regions by encoding only displacements.
However, this efficiency comes at the cost of fidelity, since the vectors are inherently coarse and contain compression artifacts stemming from their discrete block-based representation and aggressive quantization.

The RMR module is pretrained to map coarse, block-level motion vectors to fine-grained, dense motion fields.
For pretraining the RMR module, we generate dense optical flow using an off-the-shelf model (Co-Tracker3~\cite{cotracker}).
The module is then pretrained by minimizing the $L_2$ loss between its predicted motion fields and the dense optical flow targets.
This pretraining step forces the model to learn the underlying motion structures, effectively touching up the noisy inputs. 

At finetuning, we use this module as a feature encoder. It processes the raw motion vectors $\mathbf{m}^{(k, t)}$ for each inter-frame 
and outputs a sequence of motion embeddings $\bm{E}_{\text{M}}^{(k, t)} \in \mathbb{R}^{N_m \times d_s}$, where $N_m$ is the number of motion patches.
The final input for the $k$-th GOP, $\bm{Z}^{(k)}$, is then formed by concatenating the I-frame patch embeddings with all corresponding motion embeddings for that GOP:
{
\setlength{\abovedisplayskip}{6pt}
\setlength{\belowdisplayskip}{6pt}
\setlength{\abovedisplayshortskip}{6pt}
\setlength{\belowdisplayshortskip}{6pt}
\begin{align}
\label{eq:gop_input}
\bm{Z}^{(k)} = \left[ \bm{E}_{\text{I}}^{(k)}; \bm{E}_{\text{M}}^{(k, 1)}; \ldots; \bm{E}_{\text{M}}^{(k, T_g-1)} \right].
\end{align}
}
This aggregated sequence $\bm{Z}^{(k)} \in \mathbb{R}^{\text{L}_g \times d_s}$ serves as the input of the subsequent module, where $L_g = N_p + (T_g-1)N_m$.

\subsection{Hierarchical Motion State Space Module}
The core challenge in processing compressed video representation inputs is the extreme sequence length. 
Each $\text{GOP}_k$ produces a long sequence of tokens $\bm{Z}^{(k)}$.
For instance, a video with $K$ GOPs (e.g., $K=128$) results in a flat sequence of $K \times L_g$ tokens (often $> 100,000$), rendering standard attention mechanisms computationally infeasible.
To address this, we propose the HMSS module that explicitly factors the temporal reasoning into two stages, mirroring the video's natural codec structure.

\paragraph{Codec-aware Selective Scan.}
The first stage fuses motion and appearance \textit{within} each GOP.
Its goal is to create a single, motion-aware representation $\bm{h}_{\text{gop}}^{(k)}$ for each $\text{GOP}_k$.
The complete token sequence $\bm{Z}^{(k)}$ from Eq.~\eqref{eq:gop_input} is passed through a bidirectional Mamba block~\cite{Liu2024VisionMA}, which efficiently mixes the I-frame appearance embeddings $\bm{E}_{\text{I}}^{(k)}$ with their corresponding motion vector embeddings $\bm{E}_{\text{M}}^{(k, t)}$.
We then {isolate the motion-aware I-frame tokens} ${\bm{Z}}_{\text{I}}^{(k)}$ by taking the first $N_p$ tokens from the Mamba block's output:
{
\setlength{\abovedisplayskip}{6pt}
\setlength{\belowdisplayskip}{6pt}
\setlength{\abovedisplayshortskip}{6pt}
\setlength{\belowdisplayshortskip}{6pt}
\begin{align}
\label{eq:local_mamba}
    \bm{Z}_{\text{I}}^{(k)} = \text{SSM}_{\text{local}}\left(\bm{Z}^{(k)} \right)_{[1:N_p]},
\end{align}
}
where $\text{SSM}_{\text{local}}(\cdot)$ and the subscript ${[1:N_p]}$ denote the bidirectional Mamba scan and the selection of the first $N_p$ tokens from the output sequence, respectively. 
These tokens encapsulate the salient motion dynamics of the entire GOP.
See Appendix~\ref{app:ssm} for further descriptions on SSMs.

\paragraph{Bidirectional Token Mixer.}
The second stage models long-range temporal dependencies \textit{across} GOPs.
We first aggregate the summary vectors from all $K$ GOPs into a single sequence.
This sequence, which is $\text{L}_g/\text{N}_p$ times shorter than the naive flattened sequence, is processed by bidirectional Mamba layers, applying the scan method from~\cite{Islam_2025_CVPR}.
The implicit state of these layers efficiently keeps track of long-horizon dependencies across the entire clip:
{
\setlength{\abovedisplayskip}{6pt}
\setlength{\belowdisplayskip}{6pt}
\setlength{\abovedisplayshortskip}{6pt}
\setlength{\belowdisplayshortskip}{6pt}
\begin{align}
    \bm{H} = \text{SSM}_{\text{global}}\left( [\bm{Z}_{\text{l}}^{(0)}; \bm{Z}_{\text{l}}^{(1)}; \ldots; \bm{Z}_{\text{l}}^{(K-1)}] \right).
\end{align}
}

\begin{table*}[t]
    \centering
    \normalsize
    \setlength{\tabcolsep}{2.0pt}
    \begin{tabular}{lccccccc}
        \toprule
        \textbf{Method} &
        \textbf{LLM backbone} &
        \textbf{LongVideoBench} &
        \textbf{NExT-QA} &
        \textbf{MLVU} &
        \textbf{VideoMME} &
        \textbf{Perception Test} &
        \textbf{Average} \\
        \midrule
        LLaMA-VID~\cite{llamavid} & Vicuna-7B & -- & -- & 33.2 & -- & 44.6 & 38.9 \\
        Video-LLaVA~\cite{videollava} & Vicuna-7B & -- & 62.6 & 47.3 & 40.4 & 44.3 & 48.7 \\
        PLLaVA~\cite{pllava} & Vicuna-7B & 40.2 & 68.2 & 47.3 & 44.3 & -- & 50.0 \\
        VideoChat2~\cite{mvbench} & Vicuna-7B & 36.0 & -- & 54.6 & 47.9 & 47.3 & 46.5 \\
        LLaVA-NeXT-Video~\cite{llavanextvideo} & Qwen1.5-7B & 43.5 & -- & -- & 46.5 & 48.8 & 46.3 \\
        Video-LLaMA2~\cite{videollama2} & Mistral-7B & -- & 51.7 & 48.5 & 47.9 & 51.4 & 49.9 \\
        Kangaroo~\cite{kangaroo} & Llama-3-8B & -- & 62.7 & 61.0 & 56.0 & -- & 59.9 \\
        Video-LaVIT~\cite{video-lavit} & LaVIT-7B & -- & -- & -- & -- & 47.9 & 47.9 \\
        LongVA~\cite{longva} & Qwen2-7B & 41.5 & 68.3 & 56.3 & 54.3 & -- & 55.1 \\
        EMA~\cite{zhao25cvpr} & Qwen2-7B & 47.0 & -- & 57.2 & 53.4 & -- & 52.5 \\
        Video-XL~\cite{video-XL} & Qwen2-7B & 49.5 & -- & 64.9 & 55.5 & -- & 56.6 \\
        LLaVA-OneVision~\cite{llavaonevision} & Qwen2-7B & 56.5 & 79.4 & 64.7 & 58.2 & 57.1 & 63.2 \\
        LongVU~\cite{shen2025longvu} & Qwen2-7B & -- & -- & 65.4 & 60.6 & -- & 63.0 \\
        BIMBA~\cite{Islam_2025_CVPR} & Qwen2-7B & \underline{59.5} & \underline{83.2} & 70.6 & 63.1 & \textbf{68.1} & \underline{68.9} \\
        Qwen2-VL~\cite{qwen2vl} & Qwen2-7B & 55.6 & -- & -- & 63.3 & -- & 59.5 \\
        LLaVA-Video~\cite{llavavideo} & Qwen2-7B & 58.2 & \underline{83.2} & \underline{70.8} & 63.3 & \underline{67.9} & 68.7 \\
        Qwen2.5-VL~\cite{qwen25vl} & Qwen2.5-7B & \underline{59.5} & 74.6 & 70.2 & \textbf{65.1} & -- & 67.4 \\
        \rowcolor{MintGreen}
        \textbf{ReMoRa (Ours)} & Qwen2-7B & \textbf{60.8} & \textbf{84.2} & \textbf{72.1} & \underline{64.4} & 67.7 & \textbf{69.8} \\
        \bottomrule
    \end{tabular}
    \vspace{-1mm}
    \caption{Quantitative comparison of recent video MLLMs across multiple benchmarks. 
    ReMoRa achieved the best performance on LongVideoBench, NExT-QA, and MLVU, and obtains the highest overall average score of 69.8, while remaining highly competitive on VideoMME and Perception Test.
    \textbf{Bold} indicates the best performance, and \underline{underline} indicates the second best in each column.}
    \label{table:quant}
    \vspace{-2mm}
\end{table*}

The final video features $\bm{H}$ are projected into the LLM's embedding space and concatenated with the tokenized instruction embeddings $\bm{x}_{\text{txt}}$.
The LLM then generates the output token sequence $\hat{\bm{y}} = (\hat{y}_1, \hat{y}_2, \dots, \hat{y}_N)$ in an auto-regressive manner:
\begin{align}
    \hat{y}_n = \arg\max_{\tilde{y} \in \mathcal{V}} p_\theta\left(\tilde{y} \mid \bm{H}, \bm{x}_{\text{txt}}, \hat{\bm{y}}_{<n}\right),
\end{align}
where $\mathcal{V}$ is the vocabulary, $\hat{\bm{y}}_{<n}$ are the previously predicted tokens, and $\theta$ are the model parameters.
The model is trained using a standard cross-entropy loss on the predicted tokens.
Although ReMoRa incorporates several components, it does not incur substantial computational overhead, as detailed in Appendix~\ref{app:computation}.

\section{Experiments}
\label{sec:exp}

\subsection{Experimental Setup}
\paragraph{Training Data.}
For our model's training, we aggregated a comprehensive 200K instruction tuning dataset sourced from the LLaVA-Video-178K dataset~\cite{llavavideo} encompassing open-ended QA, multiple-choice QA, and captioning tasks.
These videos are sourced from a diverse collection of 10 existing datasets such as InternVid-10M~\cite{internvid}, HD-VILA-100M~\cite{hdvila}, and VIDAL~\cite{languagebind}.

We evaluate using two groups of standard benchmarks.
For long-video understanding, which targets minute to hour scale temporal reasoning, multimodal grounding, and robustness, we use LongVideoBench~\cite{longvideobench}, MLVU~\cite{mlvu}, Perception Test~\cite{perceptiontest}, VideoMME~\cite{videomme},  and NExT-QA~\cite{nextqa}.
 In addition, we used standard open-ended VideoQA benchmarks: MSVD-QA~\cite{msvdqa} and ActivityNet-QA~\cite{activitynetqa}.
For these VideoQA benchmarks, we follow the standard approach~\cite{maaz24acl}, and used GPT-3.5-turbo to evaluate the accuracy and answer quality scores.
See Appendix~\ref{app:benchmark} for further dataset descriptions and benchmark statistics.

\paragraph{Implementation Details.}
We adopted {Qwen2}~\cite{yang2024arxiv} and {SigLIP ViT-SO}~\cite{zhai23iccv} as the LLM and vision encoder backbone, respectively. 
{LoRA}~\cite{hu2022iclr} was applied to the LLM backbone and the vision encoder was kept {frozen} throughout training for parameter-efficient adaptation.

To ensure consistency across datasets and fully exploit the codec-aware design of our framework, we developed a {scene-adaptive video preprocessing pipeline}. 
All clips were spatially downsampled to $384 \times 384$ pixels and temporally resampled to 16\,fps. 
Unlike conventional uniform sampling, our approach introduces a {scene-adaptive GOP construction} based on codec keyframe detection.
Specifically, we re-encoded each video using the {H.264} codec with \textit{ffmpeg}'s scene-adaptive detection~\cite{adaptiveh264} to dynamically insert {I-frames} according to visual discontinuities. 
This procedure produces a {content-aware GOP structure} (maximum length of 32 frames), effectively performing {implicit keyframe extraction} aligned with scene transitions rather than temporal intervals. 
All frames were encoded with $4 \times 4$ macroblocks, ensuring fine-grained motion vector representation.
For further implementation details, see Appendix~\ref{app:impldetail}.

\paragraph{Baselines.}
We compare our model with a range of recent video MLLMs.
Standard video MLLMs such as LLaVA-OneVision~\cite{llavaonevision}, Qwen2-VL~\cite{yang2024arxiv}, and LLaVA-Video~\cite{llavavideo} process uniformly sampled RGB frames through vision encoders and pretrained LLMs to perform open-ended video question answering and reasoning.
Additionally, we include MLLMs that utilize codec information such as Video-LaVIT~\cite{video-lavit} and EMA~\cite{zhao25cvpr}.
Our proposed ReMoRa follows this codec-aware paradigm by incorporating motion features derived from codec streams to achieve temporally compact yet informative video representations.

\vspace{-1mm}
\subsection{Results}
\vspace{-1mm}

\subsubsection{Quantitative results}
\vspace{-1mm}
Table~\ref{table:quant} presents a quantitative comparison between ReMoRa and various baseline methods across six video understanding benchmarks. 
We report performance based on a single evaluation for all benchmarks. 
The results in Table~\ref{table:quant} indicate that ReMoRa achieved the highest scores on LongVideoBench, NExT-QA, and MLVU, with scores of 60.8, 84.2, and 72.1, respectively. 
It outperformed the second-best models in these categories by 1.3, 1.0, and 1.3 points, respectively.
Our model also demonstrated competitive performance on the VideoMME and PerceptionTest benchmarks, achieving scores of 64.4 and 67.7, respectively. The VideoMME score represents the second-best result, 0.7 points behind the top score, and the Perception Test score trails the leading results by 0.4 and 0.2 points.
Furthermore, our model's average score of 69.8 also surpasses the highest-scoring baseline models by 0.9 points. 
This consistent performance across a diverse set of tasks highlights the effectiveness of ReMoRa in long-video understanding.

Table~\ref{tab:quant2} shows the results of our evaluation on the open-ended VideoQA benchmarks MSVD-QA and ActivityNet-QA.
On the ActivityNet-QA benchmark, ReMoRa obtained the highest Accuracy and Score of 60.5 and 3.7, respectively.
This shows significant improvements, with our model outperforming the next-best model by 8.4 and 0.2 points in Accuracy and Score, respectively.
For the MSVD-QA benchmark, ReMoRa demonstrates highly competitive results, obtaining the second-best Score of 4.0, 0.1 points behind the top score, and a strong Accuracy of 73.1. 
These findings further underscore the robustness of our model in tasks requiring detailed temporal understanding and reasoning.

\begin{table}[t]
    \centering
    \normalsize
    \setlength{\tabcolsep}{5pt}
    \vspace{-0.2cm}
    \resizebox{1.0\linewidth}{!}{
    \begin{tabular}{lcccc}
        \toprule
        \multirow{2}{*}{\textbf{Method}} &
        \multicolumn{2}{c}{\textbf{MSVD-QA}} &
        \multicolumn{2}{c}{\textbf{ActivityNet-QA}} \\
        \cmidrule(lr){2-3} \cmidrule(lr){4-5}
        & \textbf{Acc.} & \textbf{Score} & \textbf{Acc.} & \textbf{Score} \\
        \midrule
        FrozenBiLM~\cite{frozenblim} & 32.2 & -- & 24.7 & -- \\
        VideoLLaMA~\cite{videollama} & 51.6 & 2.5 & 12.4 & 1.1 \\
        LLaMA-Adapter~\cite{llamaadapter} & 54.9 & 3.1 & 34.2 & 2.7 \\
        VideoChat~\cite{videochat} & 56.3 & 2.8 & 26.5 & 2.2 \\
        Video-ChatGPT~\cite{maaz24acl} & 64.9 & 3.3 & 35.2 & 2.7 \\
        BT-Adapter~\cite{btadapter} & 67.5 & 3.7 & 45.7 & 3.2 \\
        LLaMA-VID~\cite{llamavid} & 69.7 & 3.7 & 47.4 & 3.3 \\
        Chat-UniVi~\cite{chatunivi} & 65.0 & 3.6 & 45.8 & 3.2 \\
        Video-LLaVA~\cite{videollava} & 70.7 & 3.9 & 45.3 & 3.3 \\
        MovieChat~\cite{moviechat} & \underline{75.2} & 3.8 & 45.7 & 3.4 \\
        Video-LaVIT~\cite{video-lavit} & 73.2 & 3.9 & 50.1 & 3.3 \\
        EMA~\cite{zhao25cvpr} & \textbf{75.8} & \textbf{4.1} & \underline{52.1} & \underline{3.5} \\
        \rowcolor{MintGreen}
        \textbf{ReMoRa (Ours)} & 73.1 & \underline{4.0} & \textbf{60.5} & \textbf{3.7} \\
        \bottomrule
   \end{tabular}
    }
    \caption{Quantitative comparison on open-ended video question-answering benchmarks. 
    ReMoRa attains the best Accuracy and Score on ActivityNet-QA, surpassing the previous best baseline model, 
    and achieves a highly competitive Score on MSVD-QA with strong Accuracy.
    \textbf{Bold} indicates the best performance, and \underline{underline} indicate the second best. Acc. denotes the Accuracy.}
    \vspace{-4mm}
    \label{tab:quant2}
\end{table}

\subsubsection{Qualitative Results}
\begin{figure*}
    \centering
    \includegraphics[width=1.0\linewidth]{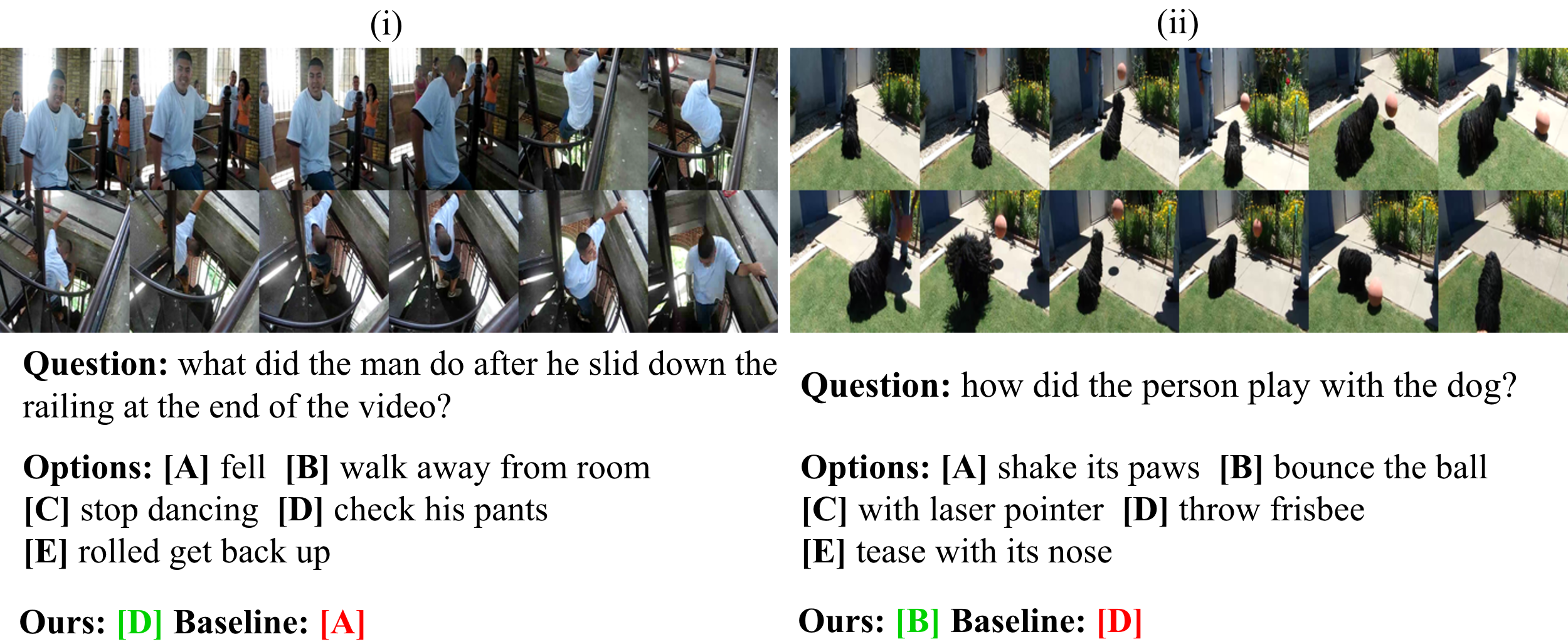}
    \caption{
    Qualitative comparison between ReMoRa and LLaVA-Video on NExT-QA.
    In both examples, ReMoRa correctly answers questions about fine-grained, temporally contextualized human actions and object motions, while LLaVA-Video fails, highlighting ReMoRa's superior use of motion cues for fine-grained action understanding.
    }
    \label{fig:qual_results_nextqa}
    \vspace{-3mm}
\end{figure*}
Fig.~\ref{fig:qual_results_nextqa} presents qualitative results from experiments on the NExT-QA dataset; in particular, it compares our model's results against those of baseline method, LLaVA-Video.
In Fig.~\ref{fig:qual_results_nextqa}, columns (i) and (ii) show two examples of the video frames, the question and answer options, our model's correct answer, and the baseline's incorrect answer, respectively.

\begin{table}[t]
  \centering
  \setlength{\tabcolsep}{2.0pt}
  \begin{tabular}{lcccc}
    \toprule
    & \textbf{\# Frames} & \textbf{Sampling} & \textbf{VideoMME} & \textbf{NExT-QA} \\
    \midrule
    \rowcolor{MintGreen}
    (a)   & 64 & CVR      & \textbf{64.3} & \textbf{84.2} \\
    (b) & 32 & CVR      & 61.9 & 82.7 \\
    (c)   & 16 & CVR      & 58.5 & 81.6 \\
    (d) & 64 & Uniform  & 62.4 & 82.8 \\
    \bottomrule
  \end{tabular}
  \small
  \caption{Ablation study on sampling and frame selection strategies.
  Using our Compressed Video Representation (CVR)-aware selection of 64 I-frames (a) achieved the best performance.
  Reducing the number of I-frames degrades results, especially on VideoMME, indicating the importance of sufficient temporal coverage.
  Simple uniform sampling (d) leads to lower scores, showing that our method is more effective than uniform sampling for long-video understanding.}
  \label{tab:abl_sampling}
  \vspace{-4mm}
\end{table}

Fig.~\ref{fig:qual_results_nextqa} (i) shows a successful example in which our model correctly answered the question ``what did the man do after he slid down the railing at the end of the video?'' by selecting option D, ``check his pants.''
This demonstrates the model's ability to capture subtle, sequential human actions.
Our approach, which utilizes motion vectors, successfully tracks the man's movement as he lands, stops, and then performs the distinct follow-up action of reaching to check his pants.
In contrast, the baseline method incorrectly chose option A, ``fell.'' This is inaccurate, as the video frames clearly show the man landing securely on his feet and remaining upright. The baseline may have misinterpreted the rapid motion of the slide and the abrupt stop as a ``fall,'' failing to process the final, stable state of the action.

Fig.~\ref{fig:qual_results_nextqa} (ii) shows another successful example in which our model, given the question ``how did the person play with the dog?'' correctly output option B, ``bounce the ball.''
This task is challenging, as observing the raw RGB frames makes it difficult to distinguish if the object is bouncing or if it is flying (e.g., being thrown horizontally).
However, the motion vectors used in our model allowed it to distinguished the specific vertical, repetitive movement pattern of a bouncing ball.
On the other hand, the baseline method inappropriately answered with option D, ``throw frisbee.''
This answer fails to capture the primary activity, as the object's trajectory is inconsistent with the horizontal, gliding motion of a frisbee.
These examples underscore our model's effective use of motion information to interpret complex actions in videos.
See Appendix~\ref{app:qualitative} for more examples.

\subsection{Ablation Studies}
We conducted ablation studies to analyze the contribution of each component in our Model (a), which corresponds to the proposed configuration described in Section~\ref{sec:method}. 
Results for each component on VideoMME and NExT-QA are summarized in Tables~\ref{tab:abl_sampling}, ~\ref{tab:abl_motion}, and ~\ref{tab:gop_strategy}.

\paragraph{Frame Selection and Temporal Resolution.}
Table~\ref{tab:abl_sampling} shows an ablation study on different frame selection and temporal sampling strategies.
Our full Model~(a) achieves the highest scores of 64.3 and 84.2 on VideoMME and NExT-QA, respectively.
Restricting the model to a maximum of 32 (b) or 16 (c) I-frames leads to a significant drop in performance.
Specifically, using the 16-frame limit (c) causes the model to underperform Model~(a) by 5.8 and 2.6 points on VideoMME and NExT-QA, respectively, demonstrating the importance of processing a sufficient number of keyframes for comprehensive understanding especially for long videos.
We also evaluated a uniform frame sampling strategy (d), which resulted in scores 1.9 and 1.4 points lower than Model~(a) on VideoMME and NExT-QA, respectively.
These results collectively highlight that our full codec-aware dynamic selection strategy is crucial to the model's performance and more effective than processing a limited number of keyframes or using simple uniform sampling.

\begin{table}[t]
  \centering
  \vspace{0.2em}
  \begin{tabular}{llcc}
    \toprule
     & \textbf{Model} & \textbf{VideoMME} & \textbf{NExT-QA} \\
    \midrule
    \rowcolor{MintGreen}
    (a) & \textbf{Full}  & \textbf{64.3} & \textbf{84.2} \\
    (f) & w/o Align         & 63.4 & 82.2 \\
    (g) & w/o RMR & 62.1 & 82.0 \\
    \bottomrule
  \end{tabular}
  \small
  \caption{Ablation study on the Refined Motion Representation (RMR) module.
  Removing optical-flow--based pretraining (f) or the RMR module (g) consistently degrades performance, showing that the RMR module with optical-flow pretraining is important for robust video understanding when using motion vectors.
  }
  \label{tab:abl_motion}
  \vspace{-3mm}
\end{table}

\paragraph{Effectiveness of the Refined Motion Representation.}
We evaluate the contributions of the RMR module and the optical-flow-based pretraining in Table~\ref{tab:abl_motion}.
Model~(f), directly finetuned without the optical-flow pretraining, underperforms Model~(a) by 0.9 and 2.0 points on VideoMME and NExT-QA, respectively, indicating that pretraining provides a useful inductive bias for interpreting motion vectors and aiding the training process.
Furthermore, Model~(g), which removes the RMR module and passes GOPs directly to the HMSS module, underperforms Model~(a) by 2.2 points on both VideoMME and NExT-QA.
This indicates that the RMR is crucial for effectively processing and utilizing noisy raw motion vectors.
Overall, the proposed Model~(a) achieves the best balance between efficiency and temporal expressiveness.

\paragraph{GOP Aggregation Strategy.}
We further ablate the design of our GOP aggregation strategy, which integrates features across I- and P/B-frames within each GOP.
As shown in Table~\ref{tab:gop_strategy}, our Model~(a) that uses the HMSS module to aggregate the GOPs achieved the highest scores of 64.3 and 84.2 on VideoMME and NExT-QA, respectively.
Replacing the module with a simple cross-attention--based fusion (h), where I-frame tokens query P/B-frame tokens and aggregate their information into the I-frame sequence, leads to a drop of 1.8 and 2.3 points on VideoMME and NExT-QA, respectively. This indicates that SSMs-based aggregation, which is tailored to modeling continuous temporal dynamics, is more effective than cross-attention for integrating motion-dominated GOP features.
Similarly, Model (i) that replaces the HMSS module with a naive additive fusion that first pools the motion features obtained from the RMR module and then adds them to the I-frame features. This results in further degradation of performance to 61.3 and 81.5 on VideoMME and NExT-QA, respectively, as it lacks a mechanism to model motion-dependent interactions within each GOP.
These results confirm that our HMSS module aggregation design effectively preserves the temporal structure encoded in GOPs, enabling richer motion--appearance integration and superior long-video reasoning.

\section{Conclusion}

\begin{table}[t]
  \centering
  \vspace{0.2em}
  \begin{tabular}{llcc}
    \toprule
     & \textbf{Model} & \textbf{VideoMME} & \textbf{NExT-QA} \\
    \midrule
    \rowcolor{MintGreen}
    (a) & \textbf{HMSS (Ours)}  & \textbf{64.3} & \textbf{84.2} \\
    (h) & Cross-attn         & 62.5 & 81.9 \\
    (i) & Add & 61.3 & 81.5 \\
    \bottomrule
  \end{tabular}
  \small
  \caption{Ablation study on the GOP aggregation strategy.
  Model (a) outperforms variants that rely on simple cross-attention (h) or naive additive fusion (i) on both benchmarks, highlighting the importance of structured temporal modeling for GOP integration.}
  \label{tab:gop_strategy}
  \vspace{-4mm}
\end{table}
In this study, the focus is on video understanding with video MLLMs. We propose ReMoRa, a video MLLM that operates directly on compressed video streams, using I-frames and motion representations in place of redundant RGB frames to achieve scalable long-form video understanding. To improve the quality of motion cues, we introduce the Refined Motion Representation module, which enhances the fidelity of noisy, block-level motion vectors, and a Hierarchical Motion State Space module that enables long-range temporal modeling with linear-time complexity. Through extensive experiments, ReMoRa is shown to outperform strong baseline methods on multiple challenging long-form video understanding benchmarks, including LongVideoBench, NExT-QA, and MLVU.

We hope that this work promotes further research on compressed-domain, motion-aware video MLLMs and helps bridge the gap between long video understanding.

\vspace{-2mm}

{\small
\bibliographystyle{ieee_fullname}
\bibliography{reference_simple}
}

\clearpage
\setcounter{page}{1}
\maketitlesupplementary

\renewcommand{\thesection}{\Alph{section}}
\setcounter{section}{0}
In this supplementary material, we provide additional background on deep state space models, an analysis of the computational efficiency of ReMoRa, dataset statistics, and further implementation details. We also present additional qualitative comparisons and error analysis, along with illustrative examples of our scene-adaptive video preprocessing and corresponding motion vectors. Video versions of these qualitative examples, together with further visualizations, are available on our project page in the supplementary materials, and we encourage readers to refer to them for a clearer sense of ReMoRa's behavior.
\section{Deep State Space Models}
\label{app:ssm}
State space models (SSMs)~\cite{gu22iclr, gu2024mamba, mamba2} have demonstrated strong capability in modeling long-range temporal dependencies while maintaining computational efficiency.
SSMs are inspired by control theory~\cite{kalman1960new}, in which a temporal process $\bm{x}(t) \in \mathbb{R} \mapsto \bm{y}(t) \in \mathbb{R}$ is represented by a $Q$-dimensional hidden state $\bm{h}(t)\in\mathbb{R}^Q$ as follows:
\begin{align}
\frac{d\bm{h}(t)}{dt} &= \mathbf{A}\bm{h}(t) + \mathbf{B}\bm{x}(t),\label{eq:state} \\
y(t) &= \mathbf{C}\bm{h}(t) + {D}\bm{x}(t),\label{eq:output}
\end{align}
where $\mathbf{A}\in\mathbb{R}^{Q\times Q}$ governs the latent dynamics, and $\mathbf{B}, \mathbf{C}, D$ are projection matrices.
By discretizing these continuous dynamics with a timescale parameter $\Delta$ and applying the zero-order hold~\cite{zhang07}, we obtain:
\begin{align}
\bm{h}_j &= \mathbf{\bar{A}}\bm{h}_{j-1} + \mathbf{\bar{B}}\bm{x}_j,\label{eq:state_zoh} \\
y_j &= \mathbf{C}\bm{h}_j + {D}\bm{x}_j.\label{eq:output_zoh}
\end{align}
Here, $\mathbf{\bar{A}} = \exp(\Delta\mathbf{A})$ and $\mathbf{\bar{B}} = (\Delta \mathbf{A})^{-1}(\exp(\Delta\mathbf{A}) - \mathbf{I})\cdot\Delta\mathbf{B}$.
While the recursive update resembles RNNs, its sequential nature hinders parallelization.
To address this, S4~\cite{gu22iclr} reformulates the system into a convolutional form:
\begin{align}
\bar{\mathbf{K}} &= \left( \mathbf{C}\bar{\mathbf{B}} + D,\ \mathbf{C}\bar{\mathbf{A}}\bar{\mathbf{B}} + D, \ldots, \mathbf{C}\bar{\mathbf{A}}^{L-1}\bar{\mathbf{B}} + D \right), \label{eq:kernel} \\
\bm{y} &= \bar{\mathbf{K}} * \bm{x}, \label{eq:conv}
\end{align}
where $L$ denotes sequence length.
This formulation enables parallel training via convolution (Eq.~\eqref{eq:conv}) and efficient inference through recurrence (Eq.~\eqref{eq:state_zoh}, ~\eqref{eq:output_zoh}).

Building on this, Mamba~\cite{gu2024mamba} introduces a dynamic selection mechanism where $\mathbf{\bar{A}}, \mathbf{\bar{B}},$ and $\mathbf{\bar{C}}$ are conditioned on the input $\bm{x}$, allowing time-varying transitions that enhance expressive capacity.
Consequently, Mamba achieves superior performance on long-sequence modeling tasks, outperforming transformers in several language and temporal domains.
Crucially, SSMs scale linearly with sequence length, offering a significant computational advantage over the quadratic complexity of standard transformers for long sequences.

Beyond generic sequence benchmarks, deep SSMs (e.g., S4, S5, and Mamba~\cite{gu22iclr,s5smith2023simplified,gu2024mamba})
have been adopted for long-sequence modeling across a wide range of domains, including robotics and video understanding~\cite{deepswm,robomamba, videomamba}.
In these settings, SSMs act as temporal backbones for models that must process very long sequences, often matching or surpassing transformer-based architectures at lower computational cost~\cite{gu22iclr,gu2024mamba,mamba2}.
However, existing video-focused SSMs~\cite{videomamba} primarily operate on dense RGB frame sequences and have not been adapted to exploit compressed-domain motion cues in multimodal video-language settings as considered in this work.

\section{Computational Efficiency}
\label{app:computation}
\begin{table}[t]
  \centering
  \setlength{\tabcolsep}{2pt}
  \begin{tabular}{lccc}
    \toprule
    \textbf{Model} & \textbf{Samples/s} & \textbf{Tokens/s} & \textbf{Max memory} \\
    \midrule
    LLaVA-Video~\cite{llavavideo} & 0.53 & 31.78 & 23.21 \\
    BIMBA~\cite{Islam_2025_CVPR}        & 0.39 & 25.16 & 10.60 \\
    ReMoRa (ours) & 0.40 & 24.45 & 10.59 \\
    \bottomrule
  \end{tabular}
  \caption{
  Throughput and peak GPU memory usage for different video MLLMs.
  ReMoRa achieves comparable samples per second and tokens per second throughput to BIMBA
  while matching its memory footprint, and it reduces peak memory usage by more than half
  compared with LLaVA-Video.
  Note that max memory is in GB.
  }
  \label{tab:efficiency}
\end{table}
We quantified the computational overhead introduced by ReMoRa in comparison with existing video MLLMs.
We report inference throughput in terms of processed video samples per second and generated tokens per second, along with the peak GPU memory usage during decoding.
For this analysis, we randomly selected 50 videos from NExT-QA and used the same subset for all models.
To ensure fairness, every model was evaluated on the same single-GPU setup (NVIDIA H200 SXM GPU) with identical video resolution, batch size, and maximum output length, and we enabled FP16 whenever the implementation allows it.

Table~\ref{tab:efficiency} captures the incremental overhead introduced by ReMoRa under these controlled conditions.
ReMoRa attains 0.40 samples/s and 24.45 tokens/s with a peak memory footprint of 10.59\,GB, closely matching the efficiency of BIMBA, while LLaVA-Video achieves 0.53 samples/s and 31.78 tokens/s at the cost of 23.21\,GB of peak memory.
These results indicate that the additional components in ReMoRa preserve computational efficiency comparable to existing strong baselines, while remaining significantly more memory-friendly than LLaVA-Video.

\section{Benchmarks}
\label{app:benchmark}

\paragraph{LongVideoBench.}
LongVideoBench~\cite{longvideobench} is a long-context video question answering benchmark constructed from 3{,}763 web videos and 6{,}678 human-written multiple-choice questions. 
It considers contexts of up to one hour and includes aligned textual signals such as subtitles, 
with a design that explicitly targets long-range temporal reasoning rather than short clip understanding.

\paragraph{MLVU.}
MLVU~\cite{mlvu} targets comprehensive long-video comprehension, spanning videos from a few minutes to nearly two hours. 
The videos are drawn from diverse sources, including movies, surveillance footage, egocentric recordings, and gameplay. 
The benchmark evaluates multiple tasks, such as open-ended question answering, multiple-choice question answering, and temporal localization, 
providing a broad view of long-form video understanding.

\paragraph{NExT-QA.}
NExT-QA~\cite{nextqa} contains 5{,}440 videos and approximately 52{,}000 manually annotated question--answer pairs. 
It is specifically designed to probe causal, temporal, and intentional reasoning about human activities, using both multiple-choice and open-ended questions to assess higher-level understanding beyond surface description.

\paragraph{VideoMME.}
VideoMME~\cite{videomme} is a large-scale evaluation suite for video multimodal large language models. 
It comprises roughly 900 videos totaling about 254 hours, with approximately 2{,}700 human-authored question--answer pairs. 
The benchmark covers short, medium, and long videos and supports multiple modalities, including visual frames, subtitles, and audio, 
enabling a comprehensive assessment of multimodal reasoning.

\paragraph{Perception Test.}
Perception Test~\cite{perceptiontest} is a diagnostic benchmark that measures core perceptual and reasoning skills such as memory, abstraction, physical reasoning, and semantic understanding. 
It uses real-world videos with dense human annotation across video, audio, and text, 
and is designed to evaluate generalization capabilities rather than simple pattern matching.

\paragraph{MSVD-QA.}
MSVD-QA~\cite{msvdqa} is derived from the MSVD captioning dataset by automatically converting captions for each short clip into question--answer pairs. 
Models are evaluated using LLM-as-a-judge, making it a standard benchmark for short video question answering.

\paragraph{ActivityNet-QA.}
ActivityNet-QA~\cite{activitynetqa} consists of around 58{,}000 human-written question--answer pairs over roughly 5{,}800 videos. 
The benchmark focuses on reasoning over complex and temporally extended web videos, requiring models to integrate information across long and diverse activities.

\section{Implementation Details.}
\label{app:impldetail}
Our model comprised approximately {7.6B} parameters and required around {2,900 T} multiply--add operations. 
Training was conducted on {16 NVIDIA H200 SXM GPUs} (141\,GB VRAM), while evaluation was performed on a single H200 GPU. 
The total training time was approximately 21 hours.
Table~\ref{tab:hyperparams} summarizes the hyperparameters used in our main experiments.
\begin{table}[t]
  \centering
  \setlength{\tabcolsep}{6pt}
  \begin{tabular}{ll}
    \toprule
    Optimizer & AdamW \\
    Learning rate & 2e-5 \\
    Batch size & 32 \\
    $T$      & $64$               \\
    $H, W$   & $384, 384$        \\
    $K$      & $64$             \\ 
    $N_m$    & $32$              \\
    $p$      & $16$               \\
    $b_h,b_w$ & $4,4$ \\
    $f_v$   & $16$ fps          \\
    \bottomrule
  \end{tabular}
  \caption{
  Hyperparameters used in our experiments.
  }
  \label{tab:hyperparams}
\end{table}
\begin{figure*}
    \centering
    \includegraphics[width=1.0\linewidth]{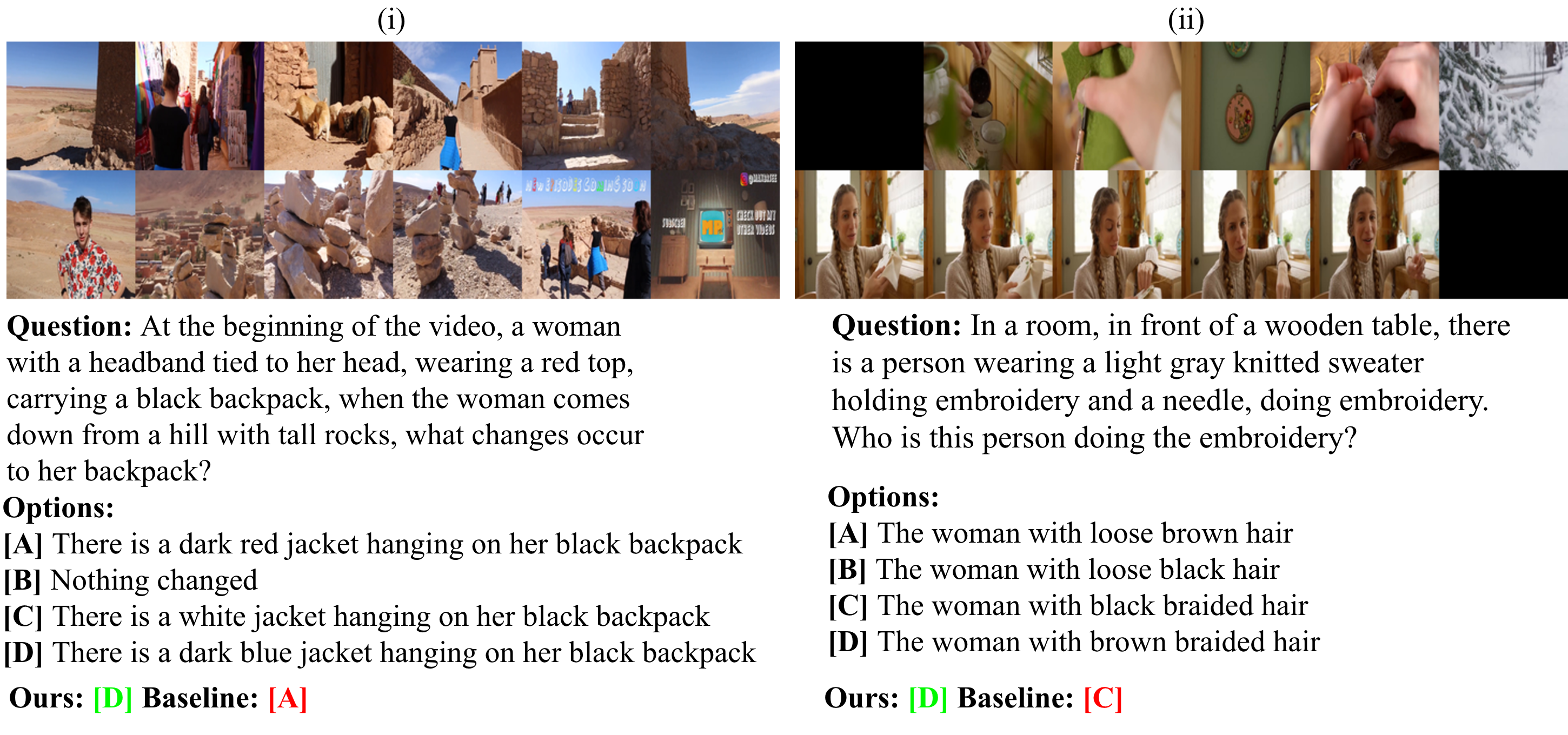}
    \caption{
Further qualitative comparison between ReMoRa and LLaVA-Video on LongVideoBench.
In both examples, ReMoRa correctly answers questions that require integrating spatial details with long-range temporal understanding, such as tracking how the scene and objects change over time and consistently identifying the person involved in the activity, while the baseline model fails.}
    \label{fig:qual_results_lvb}
    \vspace{-3mm}
\end{figure*}

\section{Additional Qualitative Results}
\label{app:qualitative}

Fig.~\ref{fig:qual_results_lvb} presents additional qualitative examples from LongVideoBench.
Panels (i) and (ii) each show sampled frames from the video, the question with multiple-choice options, and the predictions of ReMoRa and the baseline model.
In example Fig~\ref{fig:qual_results_lvb} (i), the model must track how the appearance of the backpack changes as the woman moves through the scene and descends the hill.
ReMoRa correctly reasons about the final state of the scene and selects the option describing the dark blue jacket on the backpack, whereas the baseline prediction is inconsistent with the visual evidence.

Example Fig~\ref{fig:qual_results_lvb} (ii) requires consistent identification of the person doing embroidery across multiple shots with similar backgrounds and distracting context.
ReMoRa correctly associates the description in the question with the woman who appears throughout the sequence and selects the option matching her hairstyle and appearance, while the baseline focuses on an incorrect description.

\section{Error Analysis}
\label{sec:error_analysis}

\begin{table}[t]
  \centering
  \small
  \setlength{\tabcolsep}{10pt}
  \begin{tabular}{lc}
    \toprule
    \textbf{Error mode} & \textbf{\# instances} \\
    \midrule
    Spatial comprehension error  & 22 \\
    Temporal comprehension error & 21 \\
    Motion comprehension error   & 14 \\
    Annotation error             & 10 \\
    \midrule
    \textbf{Total}               & \textbf{67} \\
    \bottomrule
  \end{tabular}
  \caption{
    Categorization of failure modes based on 67 annotated error instances from 50 randomly sampled cases where our model failed and LLaVA-Video succeeded on the NExT-QA benchmark.
    A single case may belong to multiple error modes.
  }
  \label{tab:error_analysis}
\end{table}

To better understand the limitations of ReMoRa, we conducted an error analysis.
We defined a failure case as a sample for which our model generated an incorrect answer while the baseline model (LLaVA-Video) generated the correct one.
Out of 8{,}564 evaluation samples, this criterion yielded 270 failures for our model.
From these, we randomly sampled 50 cases and manually analyzed their underlying causes.
Each case could be assigned to multiple error categories, so the total count across all error types exceeds 50, resulting in 67 annotated error instances in total.
Table~\ref{tab:error_analysis} shows the distribution over four major error modes.

\paragraph{Spatial comprehension errors.}
This category covers failures related to spatial relationships, object localization, and object presence.
Examples include incorrect reasoning about relative positions (for example, which side of the frame an object is on), confusion between nearby objects, and object hallucination where the model mentions or reasons about an object that does not appear in the video.
These cases indicate that, although keyframes provide strong appearance anchors, the current model sometimes struggles to maintain precise spatial grounding when combined with sparse motion information.

\paragraph{Temporal comprehension errors.}
Temporal errors arise when the model fails to reason over longer time spans or across multiple scenes.
Typical failure patterns include confusion about the order of events, misidentification of the stage of an activity, or inability to track how a situation evolves over time.
For instance, the model may answer a question about a later scene using information from an earlier one, or conflate two visually similar but temporally distinct segments.
This suggests that maintaining coherent temporal context across keyframes and codec-derived motion cues remains a central challenge.

\paragraph{Motion comprehension errors.}
These errors correspond to failures in understanding local motions and short-term actions.
Representative examples include questions about subtle gestures, small object manipulations, or fine-grained action transitions, where our model either misses the relevant motion cue or confuses similar actions.
Although our model employs the RMR module that transforms block-level codec motion vectors into refined motion features aligned with dense optical flow, these cases indicate that the refined signals are still not sufficiently informative for certain fine-grained dynamics.

\paragraph{Annotation errors.}
This category includes samples where the ground-truth supervision itself is unreliable.
Typical examples involve incorrect or inconsistent annotations, as well as vague questions that admit multiple plausible answers.
In such cases, our model's prediction is reasonable given the video content but is still counted as an error because it does not match the provided label.

Overall, Table~\ref{tab:error_analysis} shows that temporal and spatial comprehension errors are more frequent than pure motion comprehension errors within the subset of cases where our model fails.
Despite our model's stronger overall performance and qualitative behavior, these failure modes indicate that our model still faces challenges, not only in the quality of local motion cues, but also in how temporal and spatial features are aggregated over keyframes.

\section{Scene-aware Video Preprocessing}
\label{app:preprocessing}
\begin{figure*}
\centering
\includegraphics[width=1.0\linewidth]{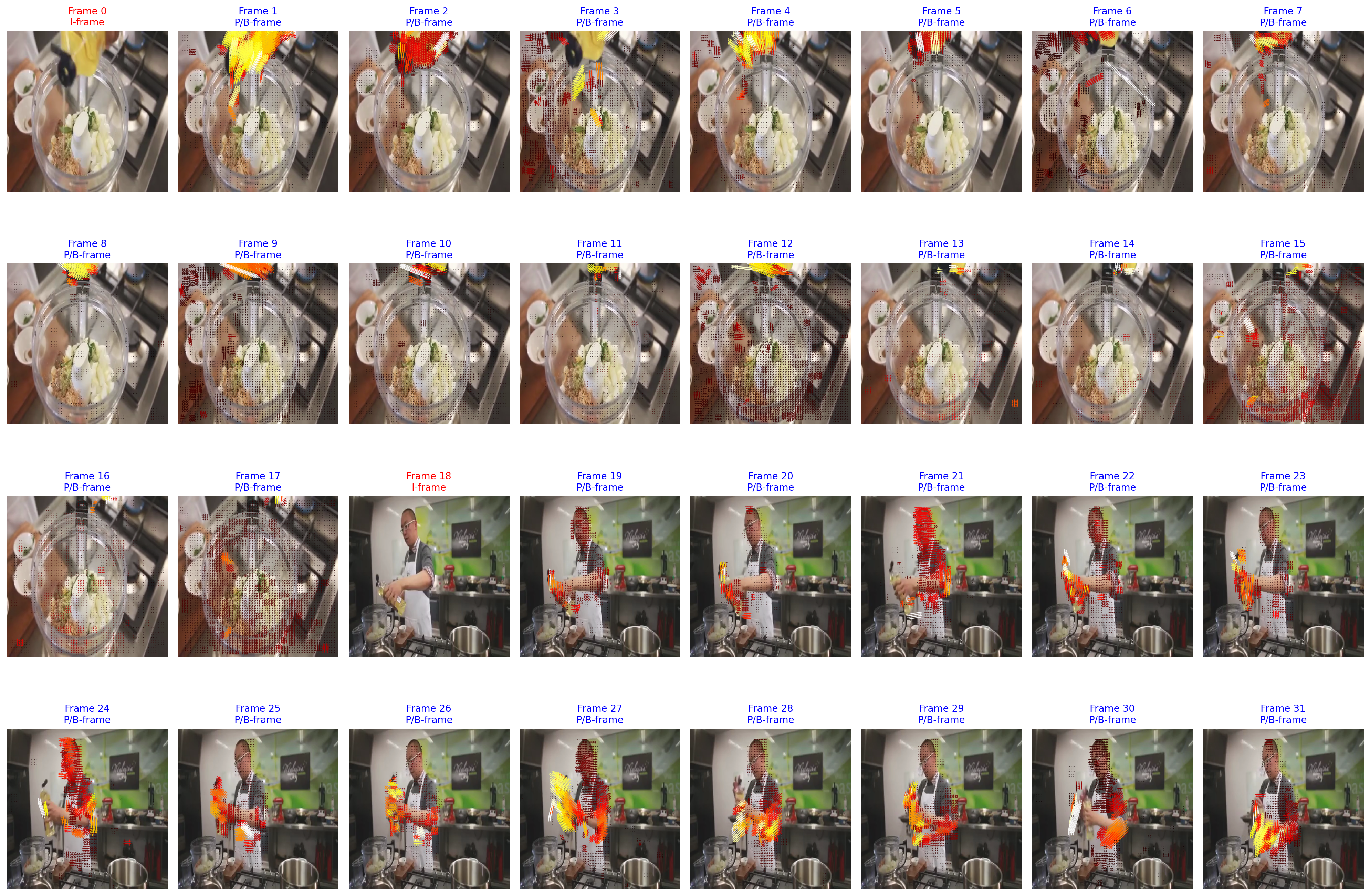}
\caption{
    Example of scene-aware video preprocessing.
    Frames 0 and 18 are scene-adaptive I-frames used as keyframes, and the remaining frames are P/B-frames with overlaid codec motion vectors.
}
\label{fig:preprocessing}
\vspace{-3mm}
\end{figure*}
Fig.~\ref{fig:preprocessing} shows an example of the proposed scene-aware video preprocessing and the corresponding motion vectors extracted from the H.264 codec.
Each row shows a sequence of frames, where scene-adaptive I-frames (e.g., Frames 0 and 18) provide appearance keyframes and the intervening P/B-frames are represented by block-wise motion vectors overlaid on the RGB images.
The sequence begins with a top-down view of a bowl and then transitions to a frontal view of the person cooking; this scene change is captured by inserting a new I-frame at Frame 9, while frames within the same scene share a stable background and exhibit locally coherent motion patterns.

The overlaid motion vectors form coarse but informative motion fields.
Large vectors concentrate around the hands, utensils, and ingredients, while the background remains mostly static.
Although codec motion vectors are block-based and sparse, they approximate the underlying optical flow by indicating the direction and magnitude of local displacements between consecutive frames.
The model interprets these vectors as pseudo optical flow and uses them to encode how objects and body parts move over time, while I-frames supply high-quality appearance cues at key timestamps.
The RMR module further refines these codec-derived motion cues by mapping them to dense optical flow targets, yielding smoother and more temporally consistent motion representations.
This preprocessing step therefore provides ReMoRa with scene-aware spatio–temporal inputs that capture both scene changes and fine-grained motions at a low computational cost.

\end{document}